# Large Language Models Don’t Make Sense of Word Problems. A Scoping Review from a Mathematics Education Perspective

Anselm R. Strohmaier[1]*, Wim Van Dooren[2], Kathrin Seßler[3], Brian Greer[4], Lieven Verschaffel[2]

[1] University of Education Freiburg, Institute for Mathematics Education
[2] KU Leuven, Centre for Instructional Psychology and Technology
[3] Technical University of Munich, Department of Educational Sciences
[4] Portland State University

*Address correspondence to: anselm.strohmaier@ph-freiburg.de

**Abstract**

The progress of Large Language Models (LLMs) like ChatGPT raises the question of how they can be integrated into education. One hope is that they can support mathematics learning, including word-problem solving. Since LLMs can handle textual input with ease, they appear well-suited for solving mathematical word problems. Yet their real competence, whether they can make sense of the real-world context, and the implications for classrooms remain unclear.

We conducted a scoping review from a mathematics-education perspective, including three parts: a technical overview, a systematic review of word problems used in research, and a state-of-the-art empirical evaluation of LLMs on mathematical word problems.

First, in the technical overview, we contrast the conceptualization of word problems and their solution processes between LLMs and students. In computer-science research this is typically labeled *mathematical reasoning*, a term that does not align with usage in mathematics education. Second, our literature review of 213 studies shows that the most popular word-problem corpora are dominated by *s-problems*, which do not require a consideration of realities of their real-world context. Finally, our evaluation of GPT-3.5-turbo, GPT-4o-mini, GPT-4.1, o3, and GPT-5 on 287 word problems shows that most recent LLMs solve these s-problems with near-perfect accuracy, including a perfect score on 20 problems from PISA. LLMs still showed weaknesses in tackling problems where the real-world context is problematic or non-sensical.

In sum, we argue based on all three aspects that LLMs have mastered a superficial solution process but do not make sense of word problems, which potentially limits their value as instructional tools in mathematics classrooms.

**Keywords**

LLM; word-problem solving; AI; mathematical reasoning; modelling

## 1 Introduction

In the last couple of years, the rapid improvement of Large Language Models (LLMs) has led to an unprecedented interest in educational research in artificial intelligence in general, and of LLMs in particular (Kasneci et al., 2023). However, while LLMs excel at producing, translating and reviewing text, they are not natively designed for processing numerical information, calculating, or proving (Chang et al., 2024). Compared to other tasks, solving mathematical problems is relatively difficult for LLMs (Testolin, 2024).

This is also true for mathematical word-problems solving. The number of studies evaluating LLMs performance on word problems is large, and several survey and review papers have recently summarized the performance of LLMs in solving word problems (Plaat et al., 2024; see also Ahn et al., 2024; Lai et al., 2024; Liu et al., 2024; Lu et al., 2023; Saraf et al., 2024; Sundaram et al., 2024; Testolin, 2024).

However, both the original studies as well as these surveys hold almost exclusively a computer science perspective and focus on LLM features and performance gains, rather than relating performance and solution processes to that of humans. Moreover, due to differing practices in how and where these studies are published, the terminology they use, and in the level of required technical knowledge to fully comprehend them, they are often not easily accessible to the mathematics education community.

From a mathematics education perspective, we don’t really know which types of word problems LLMs can solve, and which word problems they still struggle with. Moreover, while we have some knowledge about LLMs performance, there exists no comprehensive discussion of how the process of word-problem solving differs between LLMs and students. Therefore, it is unclear what the capabilities of LLMs are and how useful they are in a mathematics classroom. To address these gaps, this study provides a scoping review. It is structured in three parts, each addressing one research goal:

Part A – Conceptual foundations: We clarify from a theoretical perspective what mathematical word problems are and compare how they are solved by students and by LLMs. The goal is to lay the theoretical groundwork and to highlight similarities and differences in students' and LLM's word-problem solving.

Part B – Word-problem landscape: We provide a systematic overview of which word problems have been used in existing research on LLMs and how LLMs performance on these problems was, with the goal to investigate how these word problems align with mathematics education.

Part C – Performance analysis: We conduct our own analyses on five recent LLMs on the most used word-problem sets from computer science research, and two sets of typical and critical examples of word problems from mathematics education research. The goal is to gain a comparable and state-of-the-art understanding of the capabilities and limitations of LLMs.

Finally, we integrate these parts in a joint discussion.

## 2 Part A – Conceptual foundations: How do Students and Large Language Models Solve Word Problems?

### 2.1 Students' Word-Problem Solving

Mathematical word problems are "verbal descriptions of problem situations wherein one or more questions are raised the answer to which can be obtained by the application of mathematical operations to numerical data available in the problem situation" (Verschaffel et al., 2000, p. ix). This definition is quite broad, and a sharp division between word problems and other kinds of mathematical problems is neither possible nor helpful (Verschaffel et al., 2020). However, two key features of word problems are typical: a) the problem description includes natural language b) the problem is embedded in a real-world context. Thus, in mathematics education, word problems can be clearly distinguished from symbolic problems, which use mathematical symbols instead of natural language, or intra-mathematical problems, which are not embedded in a real-world context.

In most models of word-problem solving students are assumed to (i) build a *situation model* from the text and then (ii) translate it into a *mathematical model* for solution (Blum & Niss, 1991; Jaffe & Bolger, 2023; Kintsch & Greeno, 1985; Mayer et al., 1984; Reusser, 1990; Verschaffel et al., 2000). Thus, the solution process includes a translation from natural language to mathematics followed by mathematical analysis (Fig. 1).

Students struggle most with shifting across three representations—text, situation model, mathematical model—and with identifying the relevant information and inferring an appropriate underlying mathematical structure (Daroczy et al., 2015; Leiss et al., 2019). As a shortcut, students might suspend the connection between the problem and the real world and skip building the situation model. By applying the "rules of the game" of word-problem solving (Verschaffel et al., 2000) they directly transform the numerical information and keywords into a mathematical equation, which is also referred to as *direct translation* (Hegarty et al., 1992; Paige & Simon, 1966).

Such shortcuts often succeed in s-problems (*standard problems*), which are problems that can be solved by a straightforward application of one or more arithmetic operations with the given numbers (Verschaffel et al., 1994). However, they fail at p-problems (*problematic problems*), in which the realities of the problem context need to be taken into account in order to find an appropriate mathematical model (Verschaffel et al., 1994).

An example of a p-problem is: "Steve has bought 4 planks of 2.5 m each. How many planks of 1 m can he get out of these planks?" (Verschaffel et al., 1994, p. 276). To solve this problem correctly, it must be considered that two planks of 0.5 m can usually not be put back together to form one plank of 1m. This is different to other contexts, for example, if Steve had bought four containers of water holding 2.5 l each, he could distribute them into ten 1 l bottles.

Even more problematic is the non-sensical *age of the captain* problem: "There are 26 sheep and 10 goats on a ship. How old is the captain?", which was (incorrectly) solved with 26 + 10 = 36 years or another mathematical operation with the given numbers by 78% of first- and second-graders in the original study (IREM, 1980). Even though students might be perfectly aware that this answer makes no sense, they might believe that it is not necessary for a mathematical problem to make sense and simply follow what is likely expected in the mathematics classroom, which is to provide numerical answers problems. Schoenfeld (1991) refers to this as *suspension of sensemaking* (see also Carotenuto et al., 2021; Verschaffel et al., 2000).

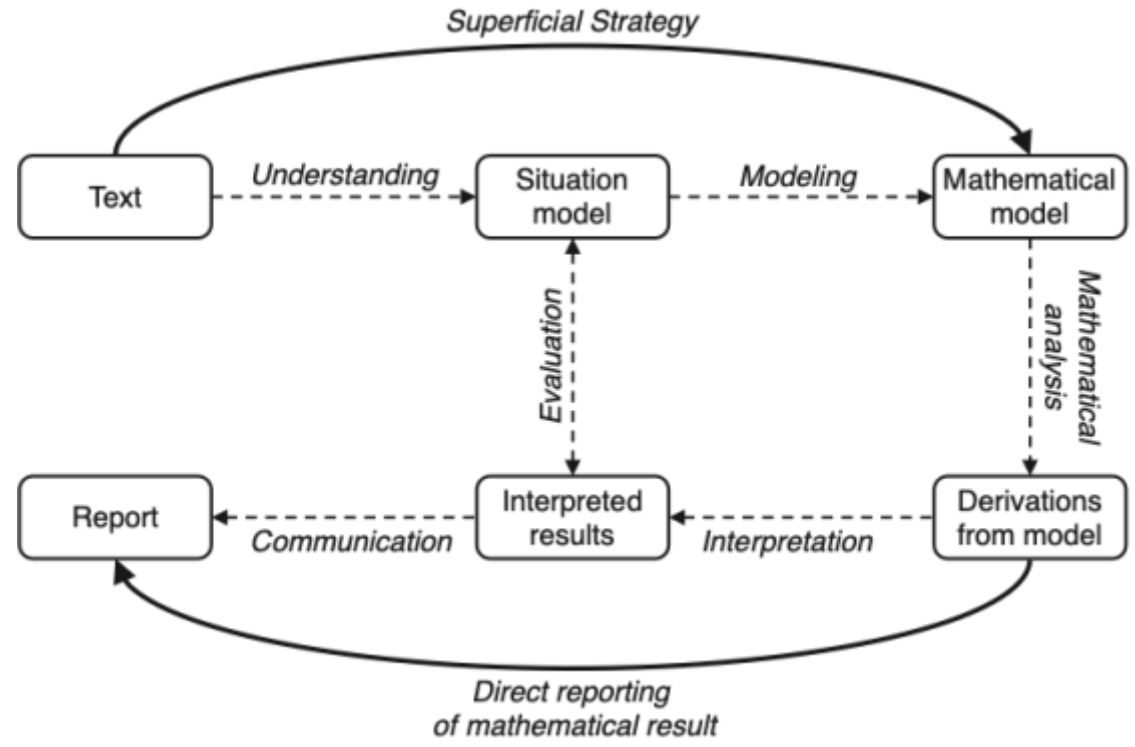

**Figure 1.** Typical and superficial solution process of mathematical word problems (Verschaffel et al., 2000, p.13)

**2.2. LLMs word-problem solving**

At first glance, chatbots such as ChatGPT appear to behave like humans. However, the underlying process differs entirely.

**2.2.1 What is a word problem for LLMs.** The previous definition of mathematical word problems is only of limited use when investigating LLMs: Any input (characters or numbers) given to an LLM is first cut down into small units, called *tokens*. This works by identifying statistically frequent sequences of characters, which are typically semantic units (e.g., Sennrich et al., 2016). Short and common words are often a single token, while longer words might be split up.

Tokenization treats numbers in the same ways as words. Essentially, this means that LLMs process mathematical problems expressed in words technically equivalent to problems expressed in mathematical symbols. LLM can still recognize the semantic difference between a mathematical symbol and a word (just like it differentiates between a verb and a noun), but in contrast to humans, a word problem is not processed in a conceptually different way than a symbolic problem. From an LLMs perspective, *every* task is a word problem. Or, since tokens are not represented as words but as vectors in a high-dimensional semantic space, all word problems are ultimately linear algebra problems. Arguably, the term *word problem* is much less meaningful and is therefore used much more liberal in computer science than in mathematics education.

**2.2.2 How does word-problem solving work for LLMs?** The previous description of how students solve word problems is also of limited use when investigating how LLMs solve them. In this article, we focus on a specific class of LLMs, which are pre-trained, decoder-only transformers (Vaswani et al., 2017). This includes models like GPT (Radford & Narasimhan, 2018) or LLaMA (Touvron et al., 2023), and in principle, Gemini (Anil et al., 2023) and Claude (Anthropic, 2024). The basic mechanism by which these LLMs generate an answer is called *autoregressive generation* (Fig. 2). When presented with a natural-language instruction (called a *prompt*), the LLM transforms the prompt into a list of tokens, determines (based on the position and meaning of the previous tokens and on pre-training data) a probability distribution for the next token and samples a token based on this distribution. The probability distribution is fully deterministic, but the sampling provides variety in the output, creating more natural sounding texts. The chosen token is then appended to the token list, step-by-step generating a text. This process repeats until either a predetermined token maximum is reached or until the *end-of-sequence* token is generated, which simply stops the process.

Consequently, when given the problem “three plus two is …” the model does not calculate 3 + 2 = 5; it has merely learned during training that the sequence “three plus two is” is most often followed by “five”. If “plus” was substituted with “minus”, this would change the probability distribution for the next token: “Five” would become unlikely, while “one” would now have a higher probability.

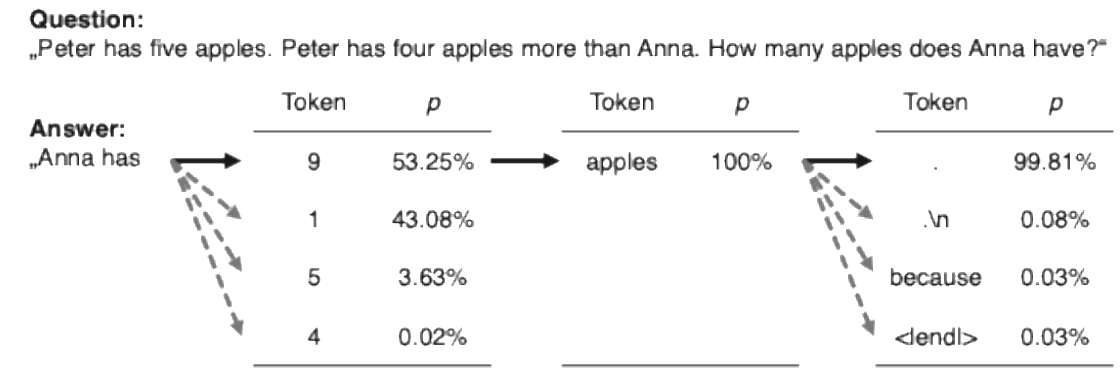

**Figure 2.** Autoregressive generation during the solution of a word problem. The figure shows the actual probability distributions for the next three tokens during an answer of GPT-3.5-turbo. *\n* is a new line, <|*end*|> is the *end-of-sequence* token. To provoke the error, this analysis was run on the older snapshot *GPT-3.5-turbo-1106*, released in November 2023. More recent models (e.g., GPT-4o-mini) solve this problem correctly.

In addition to autoregressive generation, some recent models like GPT-4o also incorporate process optimization strategies like multi-step reasoning, internal memory, or self-evaluation to improve their performance on complex tasks. They can also include external tools such as a calculator or a programming environment. These methods can substantially improve LLMs performance on word problems (Seßler et al., 2024) and arguably make the solution process more similar to that of humans: Recursive evaluation resembles the iterative steps of the modelling cycle (Blum & Leiss, 2007), though without a semantic situation model, and using external tools for calculations is similar to using a calculator. However, the main structural difference persists: LLMs will ultimately always produce text based on trained probabilities on the token level, and not by building a comprehensive situation model and deriving a mathematical model from it.

**2.2.3 Misconceptions regarding word-problem solving by LLMs.** Since LLMs appear human on the surface, some misconceptions can arise. First, LLMs cannot look back on their own process of generating text. In pure autoregressive generation, the output itself *is* the process—there is no hidden layer to analyze. Asking an LLM “How did you solve this?” simply triggers another round of autoregressive generation. Second, LLMs can only learn during pre-training and fine-tuning, but they are not self-learning systems and do not update their internal parameters during a conversation – they merely keep a limited amount of it as part of the prompt. The pre-trained state of the LLM cannot be altered by the user. Third, LLMs may seem to understand everyday situations because they can recite plausible facts, yet they lack any sense of reality (or *direct grounding*). What they “know” is only a statistical echo of their training data. Humans, by contrast, build world-models through perception and embodied experience. For instance, we learn that a balloon can be shared but not cleanly divided by watching one pop when someone tries—whereas a language model infers this only from the sparse co-occurrence patterns of "balloon" and "divide" in its training data.

## 3 Part B – Word-Problem Landscape: Which word problems are used in research on LLMs word-problem solving?

For our second research goal, we systematically investigated which sets of word problems, referred to as *word-problem corpora*, were used in studies that investigated LLM performance on word-problem solving. In computer science, these corpora are typically identified by abbreviations or acronyms and are usually called *benchmarks*. For example, GSM8k (Cobbe et al., 2021) is a collection of 8500 (hence *8k*) grade-school mathematics (*GSM*) word problems.

### 3.1 Methods (Part B)

We conducted a systematic literature search based on the PRISM statement (Moher et al., 2009). In the first step, we did a database search on SCOPUS[1], including published articles as well as preprints. Our search string included a) variants of the term *mathematical performance* and b) variants of the term *LLM*[2]. The search was conducted on January 26th, 2025, and included all publications up to 2024, resulting in 428 published papers and 633 preprints[3]. These 1061 papers were then screened in two rounds based on the following inclusion criteria: a) the paper investigated the performance of LLMs; b) the paper used mathematical tasks to evaluate performance; c) the primary research goal was to evaluate mathematical performance. We excluded paper that a) assessed mathematical theorem proving; b) did not use their own data, e.g., survey papers and systematic literature reviews.

The first round of screening was done automatically by prompting GPT-4o-mini to evaluate each abstract based on the criteria ten times independently. Papers which were excluded during all ten repetitions (383 papers) were excluded without further evaluation.

The remaining 678 papers were screened manually by the first author based on the same criteria, which led to the exclusion of 274 additional studies and 90 duplicates. For the resulting 314 studies, full texts were obtained, and it was coded by the first author which problem corpora were used in the study. 96 studies were subsequently excluded because they only used problem corpora that were not mathematical word problems in the sense of our definition, i.e., they were purely symbolic or exclusively in an intra-mathematical context. Finally, 213 studies remained. In addition, we coded the publication year and whether the study was published as a journal article, in conference proceedings, or as a preprint[4].

### 3.2 Results (Part B)

Of the 213 papers, 10 had been published as a journal article by the time of the search, 67 had been published as conference proceedings, and 136 were preprints. 4 studies were published in 2022, 39 were published in 2023, and 170 were published in 2024. There were no studies published prior to 2022.

Figure 3 shows the number of studies in which each corpus was used. It was very common that studies used several corpora ($M$ = 2.2). Overall, 84 different corpora were used.

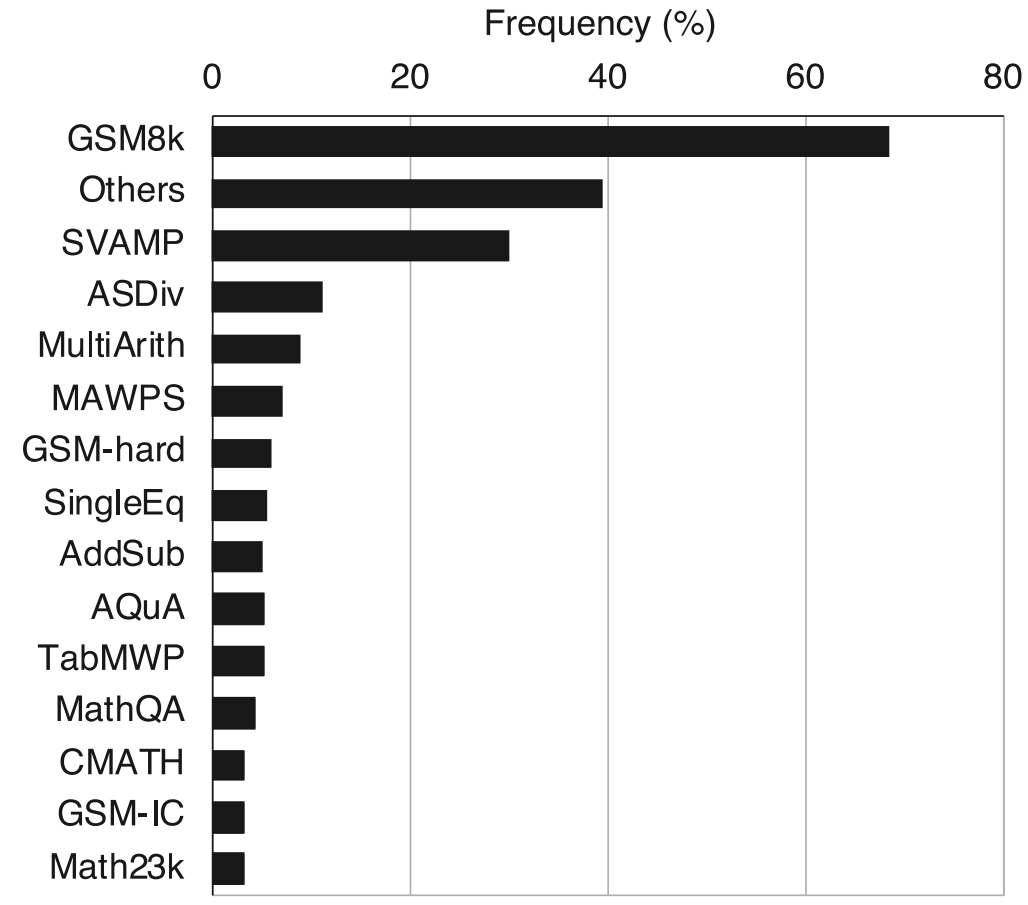

**Figure 3.** Frequencies of the word-problem corpora

### 3.3 Notable Examples

Due to the limited space, we will not discuss each of the corpora but limit ourselves to the most relevant examples. For all presented corpora, we include one example problem in Appendix A.

**3.3.1 The early challenges: MAWPS, ASDiv, and SVAMP.** One of the oldest corpora for evaluating LLMs is MAWPS (short for *Math Word Problems*, used in 15 studies; Koncel-Kedziorski et al., 2016), which is a collection of word problems from different previously published sources. Problems range from simple arithmetic to systems of two linear equations and are embedded in a real-world context. The corpus includes 3320 problems. Many studies also used one or several of the subsets of MAWPS, e.g., MultiArith, SingleEQ, or AddSub.

Although MAWPS covers a range of problem types, the structure of these problems is often quite predictable. Since AI excel at recognizing and exploiting patterns, this was considered problematic even before LLMs broke through (e.g., Zheng et al., 2022). In reaction, ASDiv (*Academia Sinica Diverse MWP Dataset*, used in 24 studies;

[1] We only used SCOPUS as this was the only available database which also includes preprints.
[2] Not including multimodal or visual models.
[3] In many cases, preprints had consecutively been published as papers. These duplicates were only removed during the coding process, since they were not always clearly distinguishable by author names and titles alone.
[4] For duplicates, we discarded the preprint, or in case of journal articles and proceedings, the proceedings.

Zheng et al., 2022) emerged as one of the most popular alternatives to MAWPS, putting a focus on providing more diverse linguistic structures. SVAMP (*Simple Variations on Arithmetic Math Word Problems*, used in 64 studies; Patel et al., 2021) is a variant of ASDiv with increased difficulty. It is the most popular ancestor of MAWPS today. SVAMP was motivated by analyses of MAWPS and ASDiv which showed that even when the question was removed, AI models still gave a correct answer to 60%-80% of all problems, illustrating their predictive and transparent structure (Patel et al., 2021).

Even though all these corpora include problems that we consider s-problems, they provided a hard challenge for AI models at the time of their release, not nearly achieving the same performance as humans. However, GPT-3 already achieved a baseline average solution rate (without any further prompting) of 73% (MAWPS), 70% (ASDiv) and 66% (SVAMP) (Wei et al., 2022), necessitating the creation of more challenging corpora.

**3.3.2 Most popular today: GSM8k.** GSM8k (Cobbe et al., 2021), was used in 146 of the 213 studies we included. It was published by a group of researchers from OpenAI. In the publication, four goals were formulated, which should address some of the issues with previous corpora: (1) The tasks should not contain errors, (2) the tasks should be linguistically diverse, so that it is not possible to solve them by identifying a template, and (3) the tasks should be of moderate difficulty, which in this case is described as “a bright middle school student should be able to solve every problem” (p. 3).

The increase in difficulty compared to earlier corpora was mostly achieved by an increase in required solution steps: GSM8k requires between 2 and 8 steps to solve, but all steps are basic arithmetic operations. From a mathematics education perspective these problems are still s-problems.

GSM8k was, at the time of its publication, quite challenging for LLMs—GPT-3’s baseline average solution rate was 16% (Wei et al., 2022) and it was described as “useful for probing the informal reasoning ability of large language models.” (Cobbe et al., 2021, p.2). This suggest that GSM8k was primarily created as a playground for training and testing the abilities of LLMs.

**3.3.3 Variants to make GSM8k more challenging**

GSM8k was a common starting point for other studies to create variant corpora for specific research questions. However, these studies mainly addressed the challenges and shortcomings of LLMs instead of increasing the difficulty that is interesting from a mathematics education perspective. For example, in *GSM-hard* (Gao et al., 2023; used in 13 studies) problem difficulty was increased by using larger numbers in the original problems, decreasing the solution rate for a GPT-3 variant from 19.7% to 5.0%. In *GSM8k-Scheherazade* (Miner et al., 2024; used in one study), up to 10 problems from GSM8k were nested, such that the quantities in one problem depended on the answer of the previous. As the context in these problems is almost never related, this resulted in quite comical (and non-sensical) problems, and decreased performance of GPT-4o from 97% to 50% for chains of length 10. In E-GSM (Xu et al., 2024; used in one study) solution rates for GPT-4o-mini decreased from around 93% to around 80% due to lengthy descriptions of the irrelevant context.

But there are also corpora that that decrease LLMs performance by increasing the relevance of the context, creating p-problems: In GSM-IC, irrelevant numerical information was added (Shi et al., 2023; used in seven studies, 80% on a GPT-3 variant), in UMP the real-world context was changed so that the question was unrealistic, but still solvable (Ma et al., 2024; used in one study, 64% on GTP-4o). NUMGlue combined mathematical problems with common sense, real-world knowledge, and reading comprehension (Mishra et al., 2022; used in five studies, 5% on GPT-3).

**3.3.4 Word-problem corpora from mathematics education research.** While most computer science corpora sets are collections of school textbook problems or are themselves generated by LLMs, we found only three studies that instead reused classic tasks from mathematics education literature—problems originally designed to probe how humans, not AI, solve word problems: Spreitzer et al. (2024) used five canonic modelling tasks (e.g., from Blum & Leiß, 2005) to investigate how LLMs perform in various subprocesses of mathematical modelling. They did not investigate solution rates systematically, but their results indicate that GPT-4.0 was able to solve most problems correctly. Schorcht et al. (2024) compared prompting techniques for mathematical problem solving, using three classical problem-based tasks in natural language (e.g., from Pólya, 1945). Similarly to Spreitzer et al. (2024), the authors did not focus on solution rates. However, they reported that LLMs solution rate was around 30% for GPT-4 and 10% for GPT-3.5 for the two harder problems. Krohling, (2023) evaluated an unknown version of ChatGPT with Bayesian reasoning problems, which were consistently solved correctly by the LLM.

## 4 Part C – Performance analysis: What is the current state of LLM’s performance in word-problem solving?

Even though the sets of tasks that were presented in the last section have been used in numerous studies, there are only little data on the state-of-the art performance of LLMs on these word problems, and no data that allows for a systematic comparison across models and corpora. This has several reasons. The first and most obvious is the rapid development of LLMs – a study that was conducted even only months ago will already be outdated. Second, since many different LLMs are available, it is hard to disentangle the performance of a specific model from the difficulty of a specific corpus.

Third, most studies use elaborate prompting techniques even for their baseline performances, with prompts often including worked examples of the same problem type. This *few-shot prompting* increases performance but does arguably not reflect a typical educational setting where teachers or students rely on models that are readily available, or where no worked example exists. Finally, most studies use automatic processes (typically utilizing LLMs) to code answers as correct or incorrect based on a numerical comparison, without considering the whole answer (Mondorf & Plank, 2024). This approach is necessary for large amounts of data, but it often limits information on the solution process.

To address these limitations, the final section reports our own evaluation: we tested five recent LLMs on four word-problem corpora without prompting and with answers coded manually.

**4.1 Methods (Part C)**

**4.1.1 Word-problem corpora** We used a random selection of 100 word problems[5] each from GSM8k (Cobbe et al., 2021) and SVAMP (Patel et al., 2021), respectively. These corpora were used since they were the most common in our analysis in Part B. The third corpus was a collection of word problems that were typical for mathematics education research: For this, we used the five modeling word problems from Spreitzer et al. (2024), the three problem-based word problems from Schorcht et al. (2024), and a collection of 20 word problems from PISA (OECD, 2006, 2013)[6]. Because of the high risk that the published PISA-problems were included in the training data for the LLMs, we also constructed 20 parallel word problems based on the same mathematical core but embedded in a different context. This led to a total of 48 problems. The fourth corpus comprised classical p-problems from four studies on sensemaking in word-problem solving (IREM, 1980; Greer, 1993; Verschaffel & De Corte, 1997; Verschaffel et al., 1994). Overall, this corpus consisted of 39 word problems.

To investigate how LLMs handle different types of word problems, we distinguished between s-problems and three types of p-problems (see Tab. 2, for examples): *Contextual problems* required real-world knowledge or assumptions to be solved correctly. These problems could be approached using the given information, but this would lead to an unrealistic answer. *Weird problems* provided all relevant information and could be solved correctly without considering the context, but the context was unrealistic. Finally, *non-sensical problems* were problems with missing information or an impossible context, making them impossible to solve meaningfully with the given information.

Initially, we expected all problems from GSM8k, SVAMP and our mathematics education corpus to be s-problems. However, during closer inspection, it turned out that of the 100 randomly selected problems from SVAMP, 9 were non-sensical problems and 17 were weird problems. For GSM8k, two problems were contextual, and three were weird (see Tab. 3).

**Table 2.** Task Types

| Task Type | Example | Source |
|---|---|---|
| Standard | If you had 4 bags with equal number of cookies and 36 cookies in total How many cookies does each bag have? | SVAMP; (Patel et al., 2021) |
| Contextual | Steve has bought 5 planks each 2,5 meters long. How many planks 1 meter long can he saw from these planks? | (Verschaffel et al., 1994) |
| Weird | There are 27 unicorns left in the world. One third of them are in the Scottish Highlands. Two thirds of the Scottish unicorns are female. How many female Scottish unicorns are there? | GSM8k; (Cobbe et al., 2021) |
| Non-Sensical | There are 26 sheep and 10 goats on a ship. How old is the captain? | (IREM, 1980) |

**Table 3.** Characteristics of task sets in the different sources

| Task Set | Sources | Size | Problem Types | | | |
|---|---|---|---|---|---|---|
| | | | Standard | Contextual | Weird | Non-sensical |
| SVAMP | (Patel et al., 2021) | 100 | 74 | 0 | 17 | 9 |
| GSM8k | (Cobbe et al., 2021) | 100 | 95 | 2 | 3 | 0 |
| Mathematics Education | (OECD, 2006, 2013; Schorcht et al., 2024; Spreitzer et al., 2024) | 48 | 43 | 5 | 0 | 0 |
| Classical p-problems | (IREM, 1980; Greer, 1993; Verschaffel & De Corte, 1997; Verschaffel et al., 1994) | 39 | 1 | 20 | 0 | 18 |

**4.1.2 LLMs.** We used five models for our evaluation. For simplicity, all five were from OpenAI: GPT 3.5 turbo (OpenAI, 2023), which is the oldest models still available as a baseline comparison, GPT 4o-mini (OpenAI, 2024a), which is the OpenAI model that is currently most affordable, as well as GPT 4.1 (OpenAI, 2025a) and o3 (OpenAI, 2025b), which were (as of June 2025) the flagship models of OpenAI. After its release in August of 2025, we added as a fifth model (OpenAI, 2025c). Although the specific technical details of the models are not fully public, GPT-4.1 is described as a more traditional chatbot based primarily on autoregressive generation, while o3 and GPT-5 are reasoning models that include several steps of reflections and chain-of-thought reasoning.

**4.1.3 Setup.** We used the official *openai* Python package (OpenAI, 2024b), which communicates with the OpenAI API to submit word problems and receives model-generated completions automatically. Each task was submitted only with the system prompt "You're solving a mathematical word problem"[7], followed by the problem text. Each of the 287 word problems were sent five times per model, resulting in a total of 1435 solutions per model, or 7175 solutions overall. In cases where no complete solution was generated due to our initial token limits (120 cases), we repeated the request with increased token limits to gain a complete dataset. For models with this parameter, temperature was set to 0.7. For GPT-5, verbosity was set to "low"; reasoning effort was set to "high".

[5] The random selection was done for cost reduction.

[6] The 20 PISA word problems contained all published items in purely textual form that had also previously been used in a PISA main study. Even though recent LLMs are capable to solve tabular or graphical problems, we limited the selection to purely textual problems for consistency.

**4.1.4 Coding.** All answers were manually coded by the first author. 8% of all answers were double coded by an independent rater to obtain an interrater reliability, κ = .84. Five answer categories were used (see Tab 4): *Wrong* answers were incorrect due to incorrect numerical data use, choosing of a wrong operation, or mistakes in calculations. *Solved* answers were solutions based on the information given in the word problem. For s-problems, this is the correct solution, while for p-problems, it is wrong. *Noticed* answers included a hint that the LLM was aware of the problematic nature of the word problem but nevertheless solved it with the given information. Answers were coded as *addressed* when the LLM made adaptions to the problem statement or made assumptions for missing information, considering aspects of the real-world to solve the problem correctly. Finally, when LLMs stated in its answer that no meaningful solution to the problem can be given (with or without a reason), it was coded as *declined*[8].

**Table 4.** Example answers for each of the five answer categories.

| Answer category | Example |
|---|---|
| Wrong | „Since the average height of the boys is 140 cm, it means half of the boys' heights are below 140 cm, and half are above 140cm." (GPT-3.5-turbo) |
| Solved | „ ... Number of weeks in the harvest season: 1181 weeks. Total money saved: 275$ x 1181 = 324 775 $. Lewis will have $324,775 at the end of the harvest season." (o3) |
| Noticed | "... Jill has -3 peaches (which is unusual, but let's continue as per the math)..." (GPT-4.1) |
| Addressed | "... Once Christmas has passed people stop buying Christmas cards, so for the three months after December you would expect the total to be about 0 cards (at most just a handful)." (o3) |
| Declined | „The captain's age is not directly related to the number of sheep and goats on the ship. The information provided does not allow us to determine the captain's age." (GPT-3.5-turbo) |

### 4.2 Results (Part C)

**4.2.1 By problem type.** The performance on s-problems was relatively high for all five models (86%-100%; see Fig. 4). o3 solved all 213 problems of this category correctly in all five repetitions. For p-problems, we observed large differences between problem types: Answers to weird problems were mostly *solved* or *wrong*, without *noticing* or *addressing* the unrealistic context. In contextual problems, the number of *addressed* answers increased from GPT-3.5-turbo (47%) to o3 (85%). An answer to non-sensical problems was *declined* in about one third of all cases across all models. This was mostly the case in classical age-of-the-captain-type problems, in which the question was clearly unrelated to the problem. In more subtle non-sensical problems, the issues were sometimes *noticed* (3% for GPT-3.5-turbo to 16% for o3) and particularly the newer models also *addressed* the issues to still provide a correct answer (24% for GPT-5). However, even o3 *solved* 29% of all non-sensical problems.

**4.2.2 By word-problem corpus.** For analyses on the level of word-problem corpora, we only distinguished between "acceptable" and "not acceptable" answers, depending on the problem type (see Tab. 5).

**Table 5.** Acceptable answers for each problem type (marked with X)

| | Acceptable answer | | | | |
|---|---|---|---|---|---|
| Problem type | Wrong | Solved | Noticed | Addressed | Declined |
| Standard | | X | | | |
| Contextual | | | | X | |
| Weird | | X | X | X | X |
| Non-sensical | | | | X | X |

Two different patterns were observed (Tab. 6): On the one hand, the rates of acceptable answers on SVAMP, GSM8k and the mathematics education corpus were quite similar: GPT-3.5-turbo solved around 80%, GPT-4o-mini solved around 90%, and GPT-4.1, o3, and GPT-5 solved around 98% of these problems with acceptable answers. Notably, o3 gave an acceptable answer in all 240 solution attempts of the 48 mathematics education word problems, including a perfect score on all PISA-problems and PISA-adaptions. On the other hand, all five models showed clear deficits in the classical p-problems. The number of acceptable answers increased substantially from GPT-3.5-turbo (51%) to GPT-5 (80%), but there is still a notable number of problems where neither model gives acceptable answers. For example, this includes the contextual "rope-problem" (Greer 1993; Appendix A.1), which was only solved correctly once, by GPT-5.

**Table 6.** Percentage of acceptable answers by word-problem corpus and LLM

| | Word-problem corpus | | | |
|---|---|---|---|---|
| LLM | SVAMP | GSM8k | Mathematics education | Classical p-problems |
| GPT-3.5-turbo | 78.6% | 82.6% | 79.2% | 51.3% |
| GPT-4o-mini | 93.2% | 95.0% | 84.6% | 56.9% |
| GPT-4.1 | 92.6% | 96.8% | 99.6% | 71.8% |
| o3 | 91.6% | 98.6% | 100.0% | 79.5% |
| GPT-5 | 92.4% | 98.0% | 97.9% | 80.0% |

[7] This prompt was used to reflect the knowledge that students might typically have when solving mathematical word problems, for example, in a mathematics classroom or as homework.

[8] Note that this is different from the LLM not providing an answer, e.g., due to reaching a token limit set by the user.

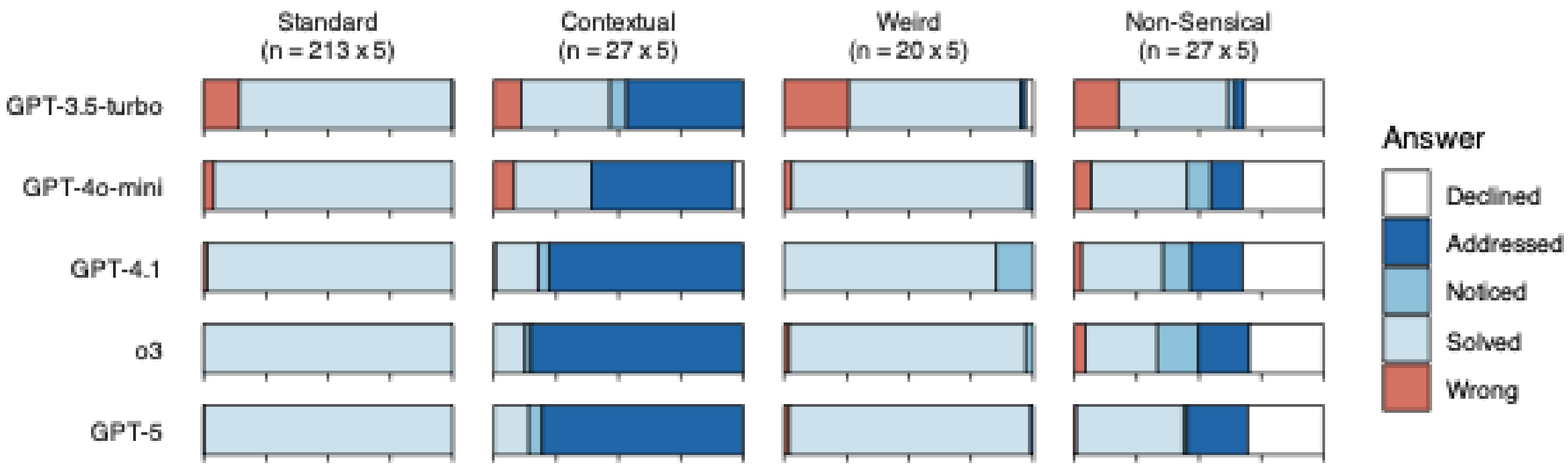

**Figure 4.** Answer categories by word-problem type for five different LLMs

## 5 Discussion

Potentially, LLMs could be a helpful tool for teaching word-problem solving in school. For this, it is important to know what the performance of LLMs on word problems is, and how their solution process theoretically relates to that of students. The present scoping review addressed this gap by integrating three perspectives: (i) an analysis of the technical mechanisms behind LLMs' word-problem solving, (ii) a literature review of the word-problem corpora used in prior studies, and (iii) a state-of-the-art performance evaluation of five contemporary models on different types of word problems. In the following, we combine these results into key findings.

### 5.1 LLMs performance is superior in s-problems, but not in p-problems

S-problems can be solved by a straightforward application of one or more arithmetic operations with the given numbers. In these word problems, LLMs perform exceptionally well. OpenAI's o3, GPT-5 and GPT-4.1 have a near-perfect performance in these problems. S-problems – especially multistep variants – can still be very challenging for students. Even though we do not have our own student data for a direct comparison, it is safe to say that LLMs have surpassed students' performance in s-problems by far. For example, our selection of word problems from PISA had an average solution rate of 41% for 15-year-olds in PISA (OECD, 2014), in contrast to the perfect score by o3.

In contrast, LLMs still make errors when solving p-problems, in which the realities of the problem context need to be considered to find an appropriate mathematical model. These errors can be further distinguished by the word-problem type: In *contextual* problems which require real-world considerations and can otherwise not be solved, LLMs were mostly capable of finding a solution. However, in *weird* and *non-sensical* problems where it appears that all relevant information is present, LLMs often fell into the trap of using only this given information instead of checking whether the problem makes sense. LLMs likely still outperform students on these challenging problems, but they do make the same errors and perform worse on p-problems than on s-problems.

It's important to note that this is a snapshot of the very recent models. However, in contrast to the speed of improvements during the last two years, the release of GPT-5 did not come with any substantial improvement in performance with regard to our analyses.

At the same time, we haven't used the models to their full potential: It seems very likely that if LLMs were trained of prompted more towards checking and including these real-world considerations, their performance in p-problems could be increased (Seßler et al., 2024).

### 5.2 LLMs don't make sense of word problems

Why do LLMs struggle with p-problems, but not with s-problems? One explanation lies in the nature of the solution process of LLMs. LLMs approach every input – whether a poem, a riddle, or a word problem – as a string of tokens. Thus, a word problem is a semantic vector space of tokens, and word-problem solving is a probabilistic linear algebra problem. What a LLM does could be described as directly translating the word problem into its solution. Thereby, they completely bypass the construction of a reality-based situation model and a mathematical model, which mathematics educators consider the very heart of mathematical modeling and sense-making. Thus, we argue that LLMs don't make sense of word problems. However, in contrast to students, LLMs are incredibly effective with this direct-translation strategy when solving s-problems. But, like students, this makes them susceptible to problems where sensemaking is essential for an adequate solution. Here efficiency becomes a liability: if the problem appears like a standard problem, the model simply works with what is given without any plausibility checks of the context. Only when the problem is not solvable otherwise, the context is included.

Another element of explanation lies in the training: Our literature review shows that the most common word-problem corpora are designed as s-problems. In part, these corpora are from times when the idea that AI might be able to solve complex word problems still seemed unthinkable, and thus, problems were purposefully designed to be relatively straightforward and de-contextualized (Kushman et al., 2014). More recent corpora however are specifically designed to target the shortcomings of LLMs to provide a challenging benchmark, but these shortcomings do not necessarily coincide with what makes word problems challenging for students. Moreover, studies from computer education typically use some form of automatic coding in their evaluation to allow for a quick and automatized evaluation of large numbers of answers. This requires the definition of unambiguous, well-defined and often numerical model solutions. This trains LLMs in a way that suggests that finding a specific solution is the goal of word-problem solving. There are only very few examples in these datasets which train towards reflecting on the problem or refusing to give an answer.

Consequently, s-problems with well-defined answers are most common for LLMs during training, and thus, their strategies might be particularly optimized for these problems. In s-problems, LLMs might just don't need other skills.

## 5.4 Theoretical and educational implications

**5.4.1 Terminology.** In computer science, the term *mathematical reasoning* is commonly used to refer to the solution of any kind of mathematical task (Sundaram et al., 2024). This is different to mathematics education in two ways: First, in computer science, the term does not distinguish between word problems and symbolic problems. This is reasonable in the context of LLMs, because both problem types are transformed into tokens in the same way. Secondly, in mathematics education the term *reasoning* is commonly reserved to activities that require some form of argumentation or logic (Hjelte et al., 2020). Solving a simple addition word problem would therefore usually not be considered reasoning. For LLMs however, the underlying process behind basic arithmetic and complex logic is the same, and thus, it is not necessary nor meaningful to make this distinction.

**5.4.2 The word-problem landscape and blind spots.** From a mathematics education perspective, the landscape of word problems used to evaluate LLMs is lob-sided towards s-problems. Even though more recent corpora tried to vary the linguistic structure and the mathematical operations, the realities of the context are still usually not relevant for their solution. In mathematics education, this is often referred to as "dressed-up" word problems. Such problems used to be common in mathematics education but nowadays have often been replaced with more contextually rich modelling problems (Verschaffel et al., 2020).

The dominance of s-problems is a problem because of the limited interpretability of many studies in an educational context, but also because models are pre-trained and optimized towards solving such problems. Many of the decisions in choosing and designing word-problem corpora come from the need to provide comparable, consistent benchmarks to evaluate the general performance of LLMs. It is not out of a research interest in human word-problem solving, but in the capabilities and limits of LLMs. For some corpora this converges (e.g., NUMGlue; Mishra et al., 2022), but for many it does not. To our knowledge there is no large word-problem corpus that is specifically curated with the requirements for genuine mathematical modelling in focus. Creating such a corpus might be a valuable task for future research.

**5.4.3 Evaluation of LLM solutions in research.** In computer science, word-problem solutions are usually evaluated automatically, often by using LLMs to compare the given answer with the correct solution. This self-containing process, which often does not involve any form of human evaluation, is very focused on the outcome rather than the solution process (Mondorf & Plank, 2024; Spreitzer et al., 2024). A more fine-grained distinction, as we tried to implement in our own manual evaluation, is more time-consuming, but seems to be much more informative from a mathematics education perspective.

**5.4.4 Relevance for mathematics classrooms.** LLMs can serve as a tool for providing solutions for s-problems. However, their use as a tutor for teaching word-problem solving is limited by the missing congruence between their own solution process and the desired process for students, as well as their own shortcomings in making sense of word problems (see also Ahn et al., 2024; Arnau-Blasco et al., 2024). If anything, LLMs can reproduce human solution processes by training, but it does not reflect their own, underlying solution.

Due to their strong performance on these s-problems, we risk over-estimating classroom readiness unless new benchmarks capture modelling, reflection and robustness. Even the most advanced models ignore issues in 30% to 40% of non-sensical problems. This seems insufficient for classroom use at least without an additional human filter. Moreover, if students interact with a chatbot while trying to solve real-world problems on their own, they might be confronted with an LLM approving their unrealistic answers (or maybe even correcting their realistic answers), similar to what was observed for teachers by Verschaffel et al. (1997).

## 5.5. Conclusion

Today, LLMs are excellent word-problem solvers. However, due to their architecture and the way they have been trained, they seem to have perfected something that students are also very good at: Solving word problems without making sense of them. Even though astonishing, this is not a desirable path for humans to follow.

# 6 References

Ahn, J., Verma, R., Lou, R., Liu, D., Zhang, R., & Yin, W. (2024, March). Large Language Models for Mathematical Reasoning: Progresses and Challenges. In N. Falk, S. Papi, & M. Zhang, *Proceedings of the 18th Conference of the European Chapter of the Association for*

*Computational Linguistics: Student Research Workshop* St. Julian's, Malta.

Anil, R., Borgeaud, S., Alayrac, J.-B., Yu, J., Soricut, R., Schalkwyk, J., Dai, A. M., Hauth, A., & Millican, K. (2023). Gemini: a family of highly capable multimodal models. *arXiv preprint arXiv:2312.11805*.

Anthropic. (2024). Claude 3 Model Card. https://www.anthropic.com/news/claude-3-family

Arnau-Blasco, J., Arevalillo-Herráez, M., Solera-Monforte, S., & Wu, Y. (2024). *Using Large Language Models to Support Teaching and Learning of Word Problem Solving in Tutoring Systems* Generative Intelligence and Intelligent Tutoring Systems: 20th International Conference, ITS 2024, Thessaloniki, Greece, June 10–13, 2024, Proceedings, Part I, Thessaloniki, Greece. https://doi.org/10.1007/978-3-031-63028-6_1

Blum, W., & Leiß, D. (2005). Modellieren im Unterricht mit der „Tanken"-Aufgabe. *mathematik lehren*, *128*, 18-21.

Blum, W., & Leiss, D. (2007). How do students and teachers deal with modelling problems. *Mathematical modelling (ICTMA 12): Education, engineering and economics*, 222-231.

Blum, W., & Niss, M. (1991). Applied mathematical problem solving, modelling, applications, and links to other subjects - state, trends and issues in mathematics instruction. *Educational Studies in Mathematics*, *22*, 37-68. https://doi.org/10.1007/BF00302716

Carotenuto, G., Di Martino, P., & Lemmi, M. (2021). Students' suspension of sense making in problem solving. *ZDM – Mathematics Education*, *53*(4), 817-830. https://doi.org/10.1007/s11858-020-01215-0

Chang, Y., Wang, X., Wang, J., Wu, Y., Yang, L., Zhu, K., Chen, H., Yi, X., Wang, C., Wang, Y., Ye, W., Zhang, Y., Chang, Y., Yu, P. S., Yang, Q., & Xie, X. (2024). A Survey on Evaluation of Large Language Models. *ACM Trans. Intell. Syst. Technol.*, *15*(3), Article 39. https://doi.org/10.1145/3641289

Cobbe, K., Kosaraju, V., Bavarian, M., Chen, M., Jun, H., Kaiser, L., Plappert, M., Tworek, J., Hilton, J., & Nakano, R. (2021). Training verifiers to solve math word problems. *arXiv preprint arXiv:2110.14168*.

Daroczy, G., Wolska, M., Meurers, W. D., & Nuerk, H. C. (2015). Word problems: A review of linguistic and numerical factors contributing to their difficulty. *Frontiers in Psychology*, *6*, Article 348. https://doi.org/10.3389/fpsyg.2015.00348

Gao, L., Madaan, A., Zhou, S., Alon, U., Liu, P., Yang, Y., Callan, J., & Neubig, G. (2023). Pal: Program-aided language models. International Conference on Machine Learning,

Greer, B. (1993). The mathematical modeling perspective on wor (l) d problems. *The Journal of Mathematical Behavior*.

Hegarty, M., Mayer, R. E., & Green, C. E. (1992). Comprehension of arithmetic word problems: Evidence from students' eye fixations. *Journal of Educational Psychology*, *84*(1), 76-84. https://doi.org/10.1037/0022-0663.87.1.18

Hjelte, A., Schindler, M., & Nilsson, P. (2020). Kinds of Mathematical Reasoning Addressed in Empirical Research in Mathematics Education: A Systematic Review. *Education Sciences*, *10*(10), 289. https://www.mdpi.com/2227-7102/10/10/289

Institut de Recherche sur l'Enseignement des Mathématiques (IREM), Équipe «Élémentaire». (1980). Quel est l'âge du capitaine? In lnstitut de Recherche sur l'Enseignement des Mathernatiques (IREM) de Grenoble (Ed.), *Bulletin de l' Association des professeurs de Mathematique de l' Enseignement Public, no 323*. https://bibnum.publimath.fr/AAA/AAA80016.pdf

Jaffe, J. B., & Bolger, D. J. (2023). Cognitive Processes, Linguistic Factors, and Arithmetic Word Problem Success: a Review of Behavioral Studies. *Educational Psychology Review*, *35*(4), 105. https://doi.org/10.1007/s10648-023-09821-6

Kasneci, E., Sessler, K., Küchemann, S., Bannert, M., Dementieva, D., Fischer, F., Gasser, U., Groh, G., Günnemann, S., Hüllermeier, E., Krusche, S., Kutyniok, G., Michaeli, T., Nerdel, C., Pfeffer, J., Poquet, O., Sailer, M., Schmidt, A., Seidel, T., . . . Kasneci, G. (2023). ChatGPT for good? On opportunities and challenges of large language models for education. *Learning and Individual Differences*, *103*, 102274. https://doi.org/https://doi.org/10.1016/j.lindif.2023.102274

Kintsch, W., & Greeno, J. G. (1985). Understanding and solving word arithmetic problems. *Psychological Review*, *92*(1), 109-129. https://doi.org/10.1037/0033-295X.92.1.109

Koncel-Kedziorski, R., Roy, S., Amini, A., Kushman, N., & Hajishirzi, H. (2016, June). MAWPS: A Math Word Problem Repository. In K. Knight, A. Nenkova, & O. Rambow, *Proceedings of the 2016 Conference of the North American Chapter of the Association for Computational Linguistics: Human Language Technologies* San Diego, California.

Krohling, R. A. (2023). Bayesian artificial brain with ChatGPT. *arXiv preprint arXiv:2308.14732*.

Kushman, N., Artzi, Y., Zettlemoyer, L., & Barzilay, R. (2014). Learning to automatically solve algebra word problems. Proceedings of the 52nd Annual Meeting of the Association for Computational Linguistics (Volume 1: Long Papers).

Lai, H., Wang, B., Liu, J., He, F., Zhang, C., Liu, H., & Chen, H. (2024). *Solving Mathematical Problems Using Large Language Models: A Survey* https://doi.org/10.2139/ssrn.5002356

Leiss, D., Plath, J., & Schwippert, K. (2019). Language and mathematics - key factors influencing the comprehension process in reality-based tasks. *Mathematical Thinking and Learning*, *21*(2), 131-153. https://doi.org/10.1080/10986065.2019.1570835

Liu, W., Hu, H., Zhou, J., Ding, Y., Li, J., Zeng, J., He, M., Chen, Q., Jiang, B., Zhou, A., & He, L. (2024). *Mathematical Language Models: A Survey* https://doi.org/10.48550/arXiv.2312.07622

Lu, P., Qiu, L., Yu, W., Welleck, S., & Chang, K.-W. (2023, July). A Survey of Deep Learning for Mathematical Reasoning. In A. Rogers, J. Boyd-Graber, & N. Okazaki, *Proceedings of the 61st Annual Meeting of the Association for Computational Linguistics (Volume 1: Long Papers)* Toronto, Canada.

Ma, J., Dai, D., Sha, L., & Sui, Z. (2024). Large language models are unconscious of unreasonability in math problems. *arXiv preprint arXiv:2403.19346*.

Mayer, R. E., Larkin, J. H., & Kadane, J. B. (1984). A Cognitive Analysis of Mathematical Problem-Solving Ability. In R. J. Sternberg (Ed.), *Advances in the Psychology of Human Intelligence* (Vol. 2, pp. 231-273). Lawrence Erlbaum Associates. https://cir.nii.ac.jp/crid/1571698599027538304

Miner, S., Takashima, Y., Han, S., Kouteili, S., Erata, F., Piskac, R., & Shapiro, S. J. (2024). Scheherazade: Evaluating chain-of-thought math reasoning in llms with chain-of-problems. *arXiv preprint arXiv:2410.00151*.

Mishra, S., Mitra, A., Varshney, N., Sachdeva, B., Clark, P., Baral, C., & Kalyan, A. (2022). NumGLUE: A suite of fundamental yet challenging mathematical reasoning tasks. *arXiv preprint arXiv:2204.05660*.

Moher, D., Liberati, A., Tetzlaff, J., & Altman, D. G. (2009). Preferred reporting items for systematic reviews and meta-analyses: the PRISMA statement. *BMJ*, *339*, b2535. https://doi.org/10.1136/bmj.b2535

Mondorf, P., & Plank, B. (2024). Beyond Accuracy: Evaluating the Reasoning Behavior of Large Language Models--A Survey. *arXiv preprint arXiv:2404.01869*.

OECD. (2006). *PISA released items - mathematics* http://www.oecd.org/pisa/38709418.pdf

OECD. (2013). *PISA 2012 released mathematics items*. https://www.oecd.org/pisa/pisaproducts/pisa2012-2006-rel-items-maths-ENG.pdf

OECD. (2014). *PISA 2012 technical report*. OECD. https://www.oecd.org/pisa/pisaproducts/PISA-2012-technical-report-final.pdf

OpenAI. (2023). GPT-3.5 Turbo fine-tuning and API updates. https://openai.com/index/gpt-3-5-turbo-fine-tuning-and-api-updates/

OpenAI. (2024a). GPT-4o mini: advancing cost-efficient intelligence. https://openai.com/index/gpt-4o-mini-advancing-cost-efficient-intelligence/

OpenAI. (2024b). OpenAI Python API library (Version 1.79.0). https://github.com/openai/openai-python

OpenAI. (2025a). Introducing GPT-4.1 in the API. https://openai.com/index/gpt-4-1/

OpenAI. (2025b). Introducing OpenAI o3 and o4-mini. https://openai.com/index/introducing-o3-and-o4-mini/

OpenAI. (2025c). Introducing GPT-5. https://openai.com/index/introducing-gpt-5/

Paige, J. M., & Simon, H. A. (1966). Cognitive processes in solving algebra word problems. In B. Kleinmuntz (Ed.), *Problem solving: Research, method, and theory* (pp. 51-119). Wiley.

Patel, A., Bhattamishra, S., & Goyal, N. (2021). Are NLP Models really able to Solve Simple Math Word Problems? North American Chapter of the Association for Computational Linguistics.

Plaat, A., Wong, A., Verberne, S., Broekens, J., van Stein, N., & Back, T. (2024). *Reasoning with large language models, a survey* https://doi.org/10.48550/arXiv.2407.11511

Pólya, G. (1945). *How to solve it*. Princeton University Press.

Radford, A., & Narasimhan, K. (2018). Improving Language Understanding by Generative Pre-Training.

Reusser, K. (1990). From text to situation to equation: cognitive simulation of understanding and solving mathematical word problems. In H. Mandl, E. De Corte, N. S. Bennett, & H. F. Friedrich (Eds.), *Learning and instruction in an international context* (pp. 477-498). Pergamon.

Saraf, A., Kamat, P., Gite, S., Kumar, S., & Kotecha, K. (2024). Towards robust automated math problem solving: a survey of statistical and deep learning approaches. *Evolutionary Intelligence*, *17*(5), 3113-3150. https://doi.org/10.1007/s12065-024-00957-0

Schoenfeld, A. H. (1991). On mathematics as sense-making: an informal attack on the unfortunate divorce of formal and informal mathematics. In J. F. Voss, D. N. Perkins, & J. W. Segal (Eds.), *Informal reasoning and education* (pp. 311-343). Erlbaum.

Schorcht, S., Buchholtz, N., & Baumanns, L. (2024). Prompt the problem–investigating the mathematics educational quality of AI-supported problem solving by comparing prompt techniques. Frontiers in Education.

Sennrich, R., Haddow, B., & Birch, A. (2016, August). Neural Machine Translation of Rare Words with Subword Units. In K. Erk & N. A. Smith, *Proceedings of the 54th Annual Meeting of the Association for Computational Linguistics (Volume 1: Long Papers)* Berlin, Germany.

Seßler, K., Rong, Y., Gözlüklü, E., & Kasneci, E. (2024). Benchmarking Large Language Models for Math Reasoning Tasks. *arXiv preprint arXiv:2408.10839*.

Shi, F., Chen, X., Misra, K., Scales, N., Dohan, D., Chi, E. H., Schärli, N., & Zhou, D. (2023). Large language models can be easily distracted by irrelevant context. International Conference on Machine Learning.

Spreitzer, C., Straser, O., Zehetmeier, S., & Maaß, K. (2024). Mathematical Modelling Abilities of Artificial Intelligence Tools: The Case of ChatGPT. *Education Sciences*, *14*(7), 698. https://www.mdpi.com/2227-7102/14/7/698

Sundaram, S. S., Gurajada, S., Padmanabhan, D., Abraham, S. S., & Fisichella, M. (2024). Does a language model “understand” high school math? A survey of deep learning based word problem solvers. *WIREs Data Mining and Knowledge Discovery*, *14*(4), e1534. https://doi.org/https://doi.org/10.1002/widm.1534

Testolin, A. (2024). Can Neural Networks Do Arithmetic? A Survey on the Elementary Numerical Skills of State-of-the-Art Deep Learning Models. *Applied Sciences*, *14*(2).

Touvron, H., Lavril, T., Izacard, G., Martinet, X., Lachaux, M.-A., Lacroix, T., Rozière, B., Goyal, N., Hambro, E., & Azhar, F. (2023). Llama: Open and efficient foundation language models. *arXiv preprint arXiv:2302.13971*.

Vaswani, A., Shazeer, N., Parmar, N., Uszkoreit, J., Jones, L., Gomez, A. N., Kaiser, Ł., & Polosukhin, I. (2017). Attention is all you need. *Advances in neural information processing systems*, *30*.

Verschaffel, L., & De Corte, E. (1997). Teaching realistic mathematical modeling in the elementary school: A teaching experiment with fifth graders. *Journal for Research in Mathematics Education*, *28*(5), 577-601.

Verschaffel, L., De Corte, E., & Borghart, I. (1997). Pre-service teachers' conceptions and beliefs about the role of real-world knowledge in mathematical modelling of school word problems. *Learning and Instruction*, *7*(4), 339-359. https://doi.org/https://doi.org/10.1016/S0959-4752(97)00008-X

Verschaffel, L., De Corte, E., & Lasure, S. (1994). Realistic considerations in mathematical modeling of school arithmetic word problems. *Learning and Instruction*, *4*(4), 273-294.

Verschaffel, L., Greer, B., & De Corte, E. (2000). *Making sense of word problems*. Swets & Zeitlinger.

Verschaffel, L., Schukajlow, S., Star, J., & Van Dooren, W. (2020). Word problems in mathematics education: A survey. *ZDM Mathematics Education*, *52*(1), 1-16. https://doi.org/10.1007/s11858-020-01130-4

Wei, J., Wang, X., Schuurmans, D., Bosma, M., Xia, F., Chi, E., Le, Q. V., & Zhou, D. (2022). Chain-of-thought prompting elicits reasoning in large language models. *Advances in neural information processing systems*, *35*, 24824-24837.

Xu, X., Xiao, T., Chao, Z., Huang, Z., Yang, C., & Wang, Y. (2024). Can LLMs Solve longer Math Word Problems Better? *arXiv preprint arXiv:2405.14804*.

Zheng, L., Long, M., Zhong, L., & Gyasi, J. F. (2022). The effectiveness of technology-facilitated personalized learning on learning achievements and learning perceptions: A meta-analysis. *Education and Information Technologies*, *27*(8), 11807-1183

## Appendix

**Figure A.1:** Example Word Problems

| Word Problem | Source |
|---|---|
| Katie picked 3 tulips and 9 roses to make flower bouquets. If she only used 10 of the flowers though, how many extra flowers did Katie pick? | MAWPS (Koncel-Kedziorski et al., 2016) |
| Sam invited 9 friends to a birthday party, but 6 couldn't come. If he wanted to buy enough cupcakes so each person could have exactly 2 how many should he buy? | ASDiv (Zheng et al., 2022) |
| Marco and his dad went strawberry picking. Together they collected strawberries that weighed 36 pounds. On the way back Marco ' dad lost 8 pounds of strawberries. Marco's strawberries now weighed 12 pounds. How much did his dad's strawberries weigh now?" | SVAMP (Patel et al., 2021) |
| Hansel makes $30,000 a year and just received a 10% raise. Gretel makes the same amount as Hansel but received a 15% raise. How much more money will Gretel make compared to Hansel? | GSM8k (Cobbe et al., 2021) |
| John wins an award at work. The award has a 1 time monetary reward of $5000 and a 5% raise in salary. If he makes 8152111 a week, every week of the year and he got the award on the first day of the year how much money did he make that year? | GSM-hard (Gao et al., 2023) |
| Emily has 4 kids named Amy, Jackson, Corey, and Jason. Amy is 5 years older than Jackson and 2 years younger than Corey. Jason is 10. Jason is 1 year younger than Corey. If it is true that Jackson is exactly 5 years old, then the following is true: [Joseph takes care of 12 dogs.] Otherwise, the following is true: [Joseph takes care of 10 dogs.] Each dog that Joseph takes care of takes .5 hours a day to walk and take care of their business. How many hours per week does Joseph spend taking care of dogs? | GSM8k-Scheherazade (Miner et al., 2024) |
| On a bright Saturday morning, a mother decided to take advantage of the weekend sales at her local supermarket. With a shopping list in hand, she navigated through the aisles, picking up items she needed for the week. Among her finds were a rich, dark cocoa powder priced at $4.20, essential for her famous chocolate cake. Next, she grabbed a bottle of laundry detergent, a necessity for the upcoming week's laundry, priced at $9.45. Lastly, she couldn't resist adding a package of pasta to her cart, a steal at just $1.35, perfect for Wednesday night's dinner. After browsing through the aisles and picking up a few more items, she made her way to the cashier. Handing over a crisp $20 bill to pay for her purchases, she waited for her change. How much change did the cashier hand back to her after her purchases? | E-GSM (Xu et al., 2024) |
| Lucy has $65 in the bank. She made a $15 deposit and then followed by a $4 withdrawal. Lucy's mother's monthly rent is $10. What is Lucy's bank balance? | GSM-IC (Shi et al., 2023) |
| Xiao Ming bought a birthday cake for his birthday. He cut the cake into 4 pieces, Xiao Hong ate 3 pieces and Xiao Li ate 2 pieces. How many pieces of cake are left? | UMP (Ma et al., 2024) |
| A man can lift one box in each of his hands. How many boxes can a group of 5 people hold in total? | NumGlue (Mishra et al., 2022) |
| Herr Stein wohnt in Tier, 20 km von der Grenze zu Luxemburg entfernt. Er fährt mit seinem VW Golf zum Tanken nach Luxemburg, wo sich direkt hinter der Grenze eine Tankstelle befindet. Dort kostet der Liter Benzin nur 0.85 Euro, im Gegensatz zu 1.1 Euro in Tier. Lohnt sich diese Fahrt für Herrn Stein? Begründe deine Antwort. | Spreitzer et al., 2024 (Blum & Leiss, 2005) |
| How can you bring up from the river exactly six quarts of water when you have only two containers, a four quart pail and a nine quart pail, to measure with? | Schorcht et al., 2024 (Polya 1945) |
| Helen has just got a new bike. It has a speedometer which sits on the handlebar. The speedometer can tell Helen the distance she travels and her average speed for a trip. Helen rode her bike from home to the river, which is 4 km away. It took her 9 minutes. She rode home using a shorter route of 3 km. This only took her 6 minutes. What was Helen's average speed, in km/h, for the trip to the river and back? | PISA (OECD, 2013) |
| A man wants to have a rope long enough to stretch between two poles 12 meters apart, but he has only one pieces of rope 1,5 meters long. How many of these pieces of rope would he need to tie together to stretch between the poles? | Greer, 1993 |
| Sven's record time at 50 meters breaststroke is 54 seconds. In how much time does he swim the 200-meter breaststroke? | Verschaffel & De Corte, 1997 |
| Grandfather gives his 4 grandchildren a box containing 18 balloons, which they share equally. How many balloons does each grandchild get? | Verschaffel et al., 1994 |
| A shepherd has 360 sheep and 10 dogs. How old is the shepherd? | IREM, 1980 |